# Neighbor Overlay-Induced Graph Attention Network

Tiqiao Wei and Ye Yuan, *Member, IEEE*

*Abstract*—Graph neural networks (GNNs) have garnered significant attention due to their ability to represent graph data. Among various GNN variants, graph attention network (GAT) stands out since it is able to dynamically learn the importance of different nodes. However, present GATs heavily rely on the smoothed node features to obtain the attention coefficients rather than graph structural information, which fails to provide crucial contextual cues for node representations. To address this issue, this study proposes a neighbor overlay-induced graph attention network (NO-GAT) with the following two-fold ideas: a) learning favorable structural information, i.e., overlaid neighbors, outside the node feature propagation process from an adjacency matrix; b) injecting the information of overlaid neighbors into the node feature propagation process to compute the attention coefficient jointly. Empirical studies on graph benchmark datasets indicate that the proposed NO-GAT consistently outperforms state-of-the-art models.

## I. INTRODUCTION

Graphs are versatile mathematical structures that build relationships between entities in a wide range of real-world situations, from blockchain networks [1, 2, 37, 38] and traffic networks [3, 39, 40, 41] to recommendation systems [4, 5, 42, 43] and protein-protein interaction networks [15, 16, 35, 44]. With the advent of Graph Neural Networks (GNNs) [8, 45, 46], there has been a notable paradigm shift in the way we approach the analysis of graph-based data [47, 48, 49, 50].

According to existing studies, GNNs have burgeoned with the develop of graph representation learning, which aims at assigning nodes in a graph to low dimensional representations and efficiently preserving the graph structures [12, 56, 57, 58]. Motivated by the success of all these numerous variants of GNNs, especially attention-based GNNs [6, 53, 54, 55], have paid great efforts to learn and represent the relationship strength based on the attention mechanism [9, 10, 51, 52]. Attention mechanism allows the GNNs to leverage different node embeddings to compute the correlations of nodes, thereby effectively capturing the local and global graph structure. Compared with other GNNs, attention-based GNNs utilize a trainable strategy to dynamically allocate weights or scores to each neighbor's information [59, 60, 71], contingent on the present state of the node representations. Through this approach, representations of nodes are well preserved, propagated across multiple layers of the network, and ultimately utilized in downstream tasks [72, 73, 81].

However, current attention-based GNNs may suffer from several inherent defects. First, the correlation between node and node is partially determined by the internal factors, i.e., layer-wise node representations in the neural network [11]. Specifically, current attention-based GNNs have not fully exploit the implicit information of the nodes, primarily due to overlooking external factors such as structural features and similarity of neighborhood nodes. Second, the attention layer, which involves transforming low dimensional node representations to higher dimensional node representations, may complicate the interpretation of the model's predictions and attention weight distributions [34, 74, 75].

Despite being effective, several approaches have been explored to ameliorate the situation [61, 62]. He et al. [11] propose conjoint attention that can combine node features with structural interventions for computing attention coefficients that can be learned outside of the GNNs for the feature aggregation. Fan et al. [17] propose an HGNN with attribute enhancement and structure-aware attention (HGNN-AESA) to tackle the issue. However, the above methods manage to obtain the weight of each node's latent position by leveraging low dimensional structural features [63, 64, 65], and thus they will not be able to capture the complete and sufficient structural features. This strategy may inevitably introduce one consequence – limited low dimensional structural properties will lead to less persuasive self-expressiveness between entire global nodes and attention coefficients for graph representation learning [79, 80, 82].

For addressing the aforementioned weaknesses, we modify and extend the idea of Neo-GNNs [18] and CATs [11] to consider both structural properties and node features in a semi-supervised attention-based GNNs way. First, we focus on exploiting the above insights to develop a new structural feature generator which is equipped with both node operator and edge operator. The scheme is aiming to acquire beneficial structural features for each node from a normalized adjacency matrix [66, 67, 84]. With such structural information, topology patterns of graphs can be easily learned and used in the subsequent tasks. Second, we discard the low-dimensional structural features generated in CATs and instead employ node operator and edge operator to construct high-dimensional structural features between overlaid neighbors, which means a neighboring node is determined based on the overlay of additional structural properties or features onto the original graph [68, 69, 70]. In this manner, the mentioned approach enables us to capture more comprehensive and detailed structural information [76, 77, 78, 83].

In this paper, we propose a simple yet efficient framework termed as Neighbor Overlay-Induced Graph Attention Network (NO-GAT) to enhance node representation learning. Specifically, our proposed model assesses potential connections among overlaid neighbors and aggregates scores from NO-GAT and feature-based GNNs in an end-to-end manner. On this basis, the attention coefficients we require are not solely determined by the input neighbor node embeddings but are also significantly influenced by external factors, such as structural properties. Theoretical analysis proves that NO-GAT have the capability to adaptively preserve feature and structural similarity. More detailed experiments reveal improvements in achieving state-of-the-art performance across a wide range of benchmark datasets.

The contributions of this work are summarized as follows:

- We propose a more practical and general semi-supervised framework, Neighbor Overlay-Induced Graph Attention Network (NO-GAT), which explicitly learns beneficial structural information from an adjacency matrix and estimates overlaid neighbors for node classification, considering both graph structural information and node features into vector representations.

- Our proposed framework can integrate structural features and node embeddings in a combined attention layer to capture appropriate node pairs neighbor similarity, i.e., attention scores, to compute proper weights for feature aggregation.

- We conduct extensive experiments on seven public benchmarks of semi-supervised node classification. Our framework consistently achieves performance improvements and the results show the effectiveness of the proposed method.

## II. RELATED WORK

**Graph Neural Networks.** GNNs are a group of neural networks that address various graph-related problem. Based on prior research, numerous spectral-based models [7] have been introduced, employing the eigenvalues and eigenvectors of the Laplacian matrix to conduct convolutions within the spectral domain. While spatial GNNs typically involve message passing schemes. The essential working process of spatial GNNs can be summarized as message-passing, aggregating and updating, which attempts to aggregate information from adjacent nodes and update graph with new information, such as GraphSAGE [13], GIN [14] and GAT [6]. These methods have been demonstrated to be able to yield positive outcomes across various tasks, including node classification [6, 7, 25], link prediction [19, 20, 21] and graph classification [22, 23].

**Semi-supervised Node Classification.** Due to the complexity of graphs in real-world application scenarios, semi-supervised learning (SSL) has been widely used in the graph field [24, 36]. SSL utilizes a small portion of labeled data and a large volume of unlabeled data to train a predictive model. For example, Scarselli et al. develop an end-to-end semi-supervised learning paradigm that was previously modeled using label propagation techniques. Besides, node classification is one of the most fundamental tasks in graphs. The goal of node classification is to classify the labeled or unlabeled nodes into a few predefined categories and ultimately achieve downstream prediction tasks. Nowadays, graph convolutional networks (GCN) [7] and graph attention networks (GAT) [6] probably are the most popular and outstanding graph neural network architecture. Typically, GCN was proposed to address semi-supervised node classification problem, where only a small quantity of the nodes have labels. And the node similarity of a central node $i$ and a neighbor $j$ is determined by the weight of their edge $A_{ij}$, which is normalized by their node degrees. However, GAT automatically acquire the importance of each neighbor nodes and assign weights to their edges. In this domain, both of them have achieved great success in the field of graph learning [85, 86].

## III. PRELIMINARIES

### A. Notations

We consider a graph $G=(V, E)$, where $V=\{v_i|i=1, 2, …, N\}$ is a set of $N$ nodes and $E$ is a set of edges between nodes. The adjacency matrix $A=\{0,1\}^{N \times N}$ is defined by $A_{ij}=1$ if $e_{ij} \in E$ and 0 otherwise. We also denote the normalized adjacency matrix $\tilde{A}=D^{-1/2}AD^{-1/2}$, where D is the diagonal degree matrix. We use $h_i \in R^F$ ($i \in \{1, 2, …, N\}$) and $h=\{h_1, h_2, …, h_N\} \in R^{N \times F}$ to represent nodes feature vectors and the collection of these vectors in the matrix form, where $F$ denotes the dimension of node features and $N$ is the number of nodes. $N_{(i)}$ denotes the directly-connected neighbors of node $i$ in the graph. $W \in R^{F' \times F}$ denotes the input linear transformation's weight matrix.

### B. Graph Attention Networks

GAT [6] employs a self-attention strategy to learn the relationships among nodes and utilizes these relationships to update the hidden features. The details of GAT propagation rules involve mining the hidden features of each node by iteratively calculating node similarities based on their features. In GAT, the final output features can be summarized as:

➢ T. Q. Wei and Y. Yuan are with the College of Computer and Information Science, Southwest University, Chongqing 400715, China (e-mail: tiqiaowei@outlook.com, yuanyekl@swu.edu.cn).

$$h'_i = \sigma(\sum_{j \in N(i)} \alpha_{ij} \mathbf{W} h_j), \tag{1}$$

Moreover, multi-head attention mechanism is raised to stabilize the learning process of self-attention. This process is formally defined as:

$$h'_i = \|_{k=1}^{K} \sigma(\sum_{j \in N(i)} \alpha_{ij}^k \mathbf{W}^k h_j), \tag{2}$$

where $\|$ represents concatenation, $\sigma(\cdot)$ represents the non-linear activation function, and $\alpha_{ij}$ are attention coefficients.

*C. Heuristic Methods*

In the context of GNNs, heuristic methods [26] can be applied to representation learning, link prediction. What makes them useful in graph learning is a result of computing the scores of node pairs similarity based on structural information [27], which can be considered as extracting predefined graph structure properties from node and edge structures of the network. One of the simplest heuristics is called Common Neighbors (CN), which measures the similarity score of a link (u, v) between two nodes by counting the number of neighbors they share:

$$f_{CN}(u,v) = |N(u) \cap N(v)|. \tag{3}$$

Another first-order heuristics is Jaccard, which measures the proportion of common neighbors:

$$f_{Jaccard}(u,v) = \frac{|N(u) \cap N(v)|}{|N(u) \cup N(v)|}. \tag{4}$$

In another perspective, Resource Allocation (RA) counts the reciprocal of the common neighbors' degrees:

$$f_{RA}(u,v) = \sum_{z \in N(u) \cap N(v)} \frac{1}{|N(z)|}. \tag{5}$$

Although heuristic methods are widely applied in the area of link prediction, these methods capture the structural information that we need in a manual way, and therefore give us good illumination to develop our model, which is capable of generalizing heuristic methods to handle overlaid neighbors for node classification in graphs.

## IV. METHODS

In this section, we elaborate on our NO-GAT. Figure. 1 shows the overall architecture of our model. In the following, we will focus on the construction of structural properties and key idea of the combined attention layer in neighbor overlay-induced graph attention network.

*A. Construction of Structural Properties*

Previous heuristic methods for capturing structural information usually rely on the immutable adjacency matrix. It is known that the whole process of these methods completely ignores the weights of connections between nodes, solely focusing on simple topological features. And each heuristic method relies on manually designed structural features extracted from node neighborhoods, which is not appropriate enough in time and efficiency.

Thus, what we want is to dynamically and automatically learn the node pair similarity scores. To achieve this aim and fully capture the usable information in structural features while mitigating potential disturbance from the graph structure, we employ structural feature generator $\Phi_\varphi(\cdot)$ comprised of two multi-layer perceptron (MLP), $f_{node}$ and $f_{edge}$, to model these features, which generate structural features of each node using an only adjacency matrix $\tilde{A} \in R^{N \times N}$ of the graph as:

$$h_i^{struct} = F_\varphi(\tilde{A}_i) = f_{node}\left(\sum_{j \in N(i)} f_{edge}(\tilde{A}_{ij})\right), \tag{6}$$

where $h_i^{struct} \in R^{N \times 1}$ contains the value of the structural features of all neighbors of node $i$. In this way, $\Phi_\varphi(\cdot)$ is applying these two MLPs with the ReLU activation function to learn structural features by capturing one-hop neighborhood information.

Once we finish constructing structural features of each node, we consider modeling overlaid neighbors' aggregation strategy. To make it solvable, we put forward the following process. First, we use the structural feature vector $h^{struct} \in R^{N \times 1}$ to create a diagonal matrix $H^{struct} \in R^{N \times N}$ as:

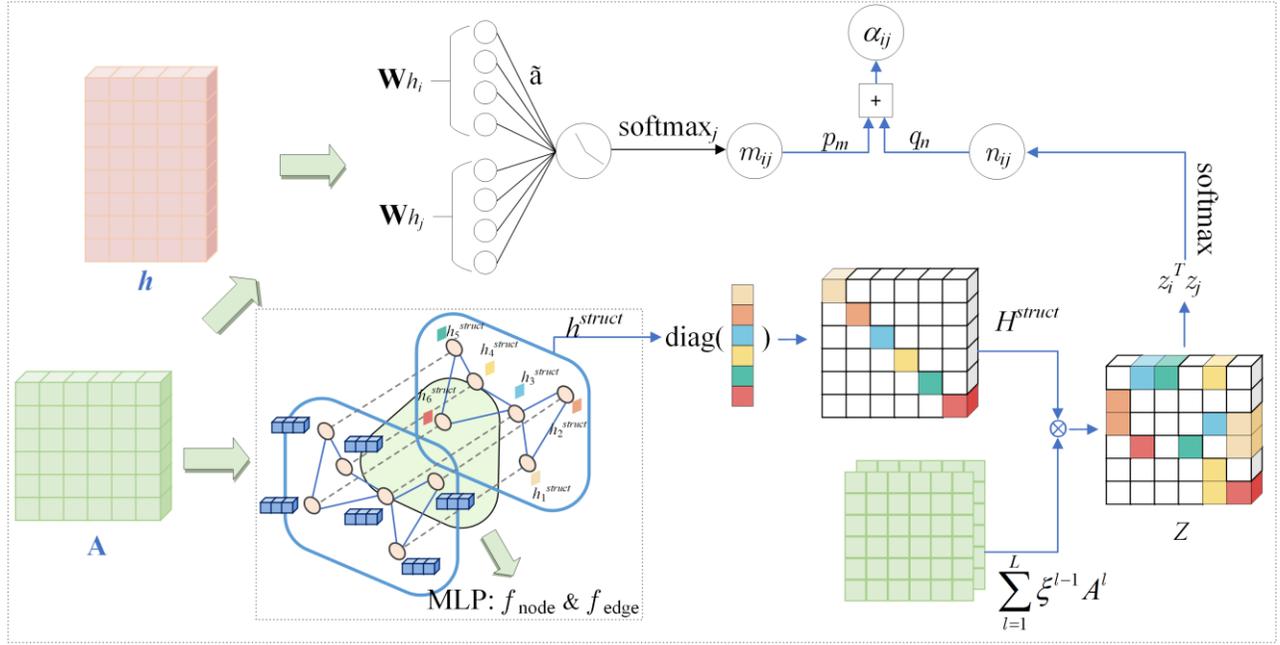

Figure 1. The NO-GAT framework for node classification

$$H^{struct} = \text{diag}(h^{struct}). \quad (7)$$

Then, taking the number of overlaid neighbors into consideration, the aggregation strategy is multiplying a normalized adjacency matrix A as:

$$Z = \tilde{A} H^{struct}. \quad (8)$$

Theoretically, the reason why we use the normalized adjacency matrix $\tilde{A}$ rather than $A$ is that $\tilde{A}$ is able to preserve neighbor importance and extract the normalized neighbors uniformly, which can hugely elevate the accuracy of the prediction and achieve promising performance, thereby preserving richer structural information in neighborhood. To take account of multi-hop overlaid neighbors, we have:

$$Z = G_\theta ( \sum_{l=1}^{L} \xi^{l-1} A^l H^{struct} ), \quad (9)$$

where $\Gamma_\theta$ is a MLP that controls the scale of $Z$ and $\xi$ represents a hyper-parameter that determines the relative significance of nearby neighbors compared to distant ones.

In this way, $z_i$ denotes each $i$-th row of the resulting $Z$ matrix, which contains the whole structural features of node $i$'s neighborhood nodes. That is to say, if we randomly select any two row vectors, e.g., $z_i$ and $z_j$, from the $Z$ matrix and perform an inner product operation, we can compute the similarity score of overlaid neighbors between the corresponding nodes. And $C_{ij}$ tries to acquire the structural correlation between neighbor nodes. It can be calculated with the following formula:

$$\mathbf{C}_{ij} = z_i^T z_j, \quad (10)$$

where $\cdot^T$ presents transposition. Having obtained the similarity score, we can describe the strong and weak relationship between two neighbor nodes. Specifically, $\mathbf{C}_{ij}$ with higher value implies that the graph structure of node $i$ can be more accurately represented by that of node $j$ compared with other nodes in the graph, indicating a stronger structural correlation between these two neighbor nodes.

*B. Neighbor Overlay-Induced Graph Attention Network*

Neighbor nodes are of great importance for the target node to learn representations. However, previous attention-based GNNs solely focus on the node embeddings inside the neural net, ignoring the fact that the original graph structure still contains lots of useful information, and it can be calculated by Eq. (10). So that we choose to use a generic graph attention mechanism, which can take structural properties, such as neighbor nodes relevance, into consideration. Specifically, adaptively combining the structural features and node embeddings arrives at an effective method to compute attention coefficients. Therefore, in NO-

GAT, we employ combined attention layers, which are more sensitive to those neighbors with the similar embeddings in layer-wise aggregation.

The input to combined layer is a set of node features, $h=\{h_1, h_2, …, h_N\} \in R^{N \times F}$. Based on the relevance of node feature embeddings and structural properties, the combined layer generates a new set of node features, $h'=\{h'_1, h'_2, …, h'_N\} \in R^{N \times F'}$ as its output. In this manner, we can aggregate neighboring nodes with weights from attention mechanism raised by graph attention networks (GAT) [6]:

$$m_{ij} = \frac{\exp(\text{LeakyReLU}(\tilde{a}^T(Wh_i \| Wh_j)))}{\sum_{k \in N(i)} \exp(\text{LeakyReLU}(\tilde{a}^T(Wh_i \| Wh_k)))}, \quad (11)$$

where $\tilde{a} \in R^{2F'}$ is a feedforward layer in neural network, $\|$ stands for the concatenation operation, and $W \in R^{F' \times F}$ is a weight matrix initialized by attention mechanism. Having obtained the normalized attention coefficients, node features that inside of graph have been successfully leveraged by GAT.

Furthermore, recalling the potential overfitting deficiency caused by multi-head attention, the bottleneck probably is that there exists limited node-node relevance, which is external to the neural net. To address this problem, the combined layer endeavor to capture structural correlation. Therefore, given structural correlation $C_{ij}$ between two neighbor nodes, the combined layer normalizes $C_{ij}$ using the softmax function to seek for additional correlation, obtaining structure coefficient $n_{ij}$:

$$n_{ij} = \text{softmax}(C_{ij}) = \frac{\exp(C_{ij})}{\sum_{k \in N(i)} \exp(C_{ik})}. \quad (12)$$

Now, we intend to compute the attention scores with respect to node-node relevance. And each combined layer can automatically acquire their relative significance. To achieve this, the combined layers bring in two learnable parameters, $g_m$ and $g_n$, respectively, to assess the relative significance of node feature and structural correlations as:

$$p_m = \frac{\exp(g_m)}{\exp(g_n) + \exp(g_m)}, q_n = \frac{\exp(g_n)}{\exp(g_n) + \exp(g_m)}, \quad (13)$$

where $p_m$ or $q_n$ refer to the normalized structural significance, which can be back propagated and dynamically updated. Unlike previous works, which directly compute attention coefficients in a feature-based attention mechanism, we combine the attention coefficient $m_{ij}$ and the structure coefficient $n_{ij}$ to obtain a combined coefficient $\alpha_{ij}$. Thus, $\alpha_{ij}$ provides an alternative way to capture the normalized weighted attention considering both graph structural information and node features. On this basis, the combined coefficient $\alpha_{ij}$ calculated by the attention mechanism can be finally expressed as:

$$\alpha_{ij} = \frac{p_m \cdot m_{ij} + q_n \cdot n_{ij}}{\sum_{k \in N(i)} \left[ p_m \cdot m_{ik} + q_n \cdot n_{ik} \right]} = p_m \cdot m_{ij} + q_n \cdot n_{ij}. \quad (14)$$

Formally, multi-head attention scheme is adopted to improve the representational learning capacity of NO-GAT. Once obtained, the combined coefficient is utilized to compute a linear combination of the features to obtain final output vector representation as:

$$h'_i = (\alpha_{ii} + \varepsilon \cdot \frac{1}{|N(i)|})Wh_i + \sum_{j \in N(i), j \neq i} \alpha_{ij} Wh_j, \quad (15)$$

where $\varepsilon$ is a learnable parameter and $W$ is the linear transformation weight matrix to estimate by backpropagation.

After acquiring the output by Eq. (15), we can eventually compute the loss function $L$. Since structural correlation $C_{ij}$ is originated from MLPs, the weights in MLPs are learnable and they can be updated after the backpropagation process. Thus, the loss function $L$ of NO-GAT is defined as follows:

$$L = L_{predict} + \lambda \|\Theta\|^2, \quad (16)$$

where $L_{predict}$ denotes the down-stream task loss for node classification, e.g., negative log-likelihood loss or cross entropy loss, $\Theta$ denotes the learnable weight parameters, $\lambda$ serves as regularization coefficient.

## V. Experiments and Analysis

In this section, we perform extensive experiments to evaluate the performance of proposed NO-GAT on node classification benchmarks.

A. *Experiments Settings*

*Datasets:* For comparison, we use seven publicly available real-world datasets of different scales for model evaluation. They are two citation networks Cora, Citeseer, one Wikipedia networks Squirrel, and three webpage networks Texas, Cornell and Wisconsin[1]. More detailed characteristics of the datasets can be found in Table 1.

TABLE 1: DATASETS STATISTICS

| Datasets | #Nodes | #Edges | #Features | #Classes |
|---|---|---|---|---|
| Cora | 2,708 | 10,556 | 1,433 | 7 |
| Citeseer | 3,327 | 9,228 | 3,703 | 6 |
| Texas | 183 | 325 | 1,703 | 5 |
| Cornell | 183 | 298 | 1,703 | 5 |
| Wisconsin | 251 | 515 | 1,703 | 5 |
| Squirrel | 5,201 | 217,073 | 2,089 | 5 |
| Actor | 7600 | 33,391 | 931 | 5 |

*Baselines:* To demonstrate the effectiveness of the proposed framework in node classification, we choose the following representative semi-supervised learning baselines including the state-of-the-art GNN models, including GCN [7], GAT [6], SGC [28], GeomGCN [29], CAT-I [11], PTDNet [30], FAGCN [31], DFI-GCN [32] and D$^2$PT [33].

*Implementation Details:* Implementation details are demonstrated in Table 2, where lr stands for learning rate.

TABLE 2: KEY SETTINGS OF DIFFERENT APPROACHES

| Datasets | $\lambda$ | lr | hidden heads | hidden layer | dropout |
|---|---|---|---|---|---|
| Cora | 0.01 | 0.005 | 4 | 8 | 0.5 |
| Citeseer | 0.01 | 0.005 | 8 | 8 | 0.5 |
| Texas | 0.001 | 0.005 | 4 | 16 | 0.5 |
| Cornell | 0.001 | 0.005 | 8 | 8 | 0.5 |
| Wisconsin | 0.001 | 0.005 | 4 | 8 | 0.5 |
| Squirrel | 0.1 | 0.005 | 8 | 128 | 0.5 |
| Actor | 0.001 | 0.005 | 8 | 128 | 0.5 |

B. *Results on Node Classification*

Results are summarized in Table 3 where we implement the semi-supervised node classification task and compare the accuracy of node classification on seven datasets. As shown in Table 3, we can observe that NO-GAT consistently outperforms all other methods across all datasets. Especially, Neo-GNNs show positive improvements on Texas and Wisconsin, where the improvements over the best baseline are 1.1% and 1.9%, respectively. Moreover, there is more than 3% improvement in classification accuracy on Squirrel, proving our proposed learning strategy that combine structural information and node features in an automatic way is effective. Besides, NO-GAT has achieved significant improvements on small datasets, including Texas, Cornell, and Wisconsin. This is attributed to the fact that small datasets are more susceptible to over-smoothing issues compared to large datasets.

TABLE 3: COMPARISONS ON NODE CLASSIFICATION (PERCENT). THE BEST RESULTS ARE IN BOLD

| Datasets | Cora | Citeseer | Texas | Cornell | Wisconsin | Squirrel | Actor |
|---|---|---|---|---|---|---|---|
| GCN | 80.85 | 70.88 | 49.57 | 51.18 | 38.47 | 35.71 | 21.38 |
| GAT | 81.08 | 70.01 | 48.65 | 44.68 | 39.61 | 31.52 | 20.74 |
| SGC | 80.70 | 68.00 | 59.46 | 58.97 | 62.75 | 30.74 | 34.14 |
| GeomGCN | 84.93 | 73.01 | 57.52 | 56.52 | 53.96 | 38.13 | 27.55 |
| CAT-I | 84.80 | 72.80 | 67.57 | 61.54 | 70.59 | 30.55 | 31.05 |
| PTDNet | 82.80 | 68.90 | 54.05 | 56.41 | 84.31 | 27.37 | 35.52 |
| FAGCN | 84.60 | 72.80 | 73.68 | 63.15 | 80.76 | 38.38 | 34.34 |
| DFI-GCN | 79.40 | 71.80 | 75.68 | 66.67 | 74.51 | 23.34 | 36.39 |
| D$^2$PT | 81.33 | 66.90 | 77.21 | 76.53 | 66.91 | 31.27 | 35.65 |
| **NO-GAT** | **85.20** | **73.10** | **78.38** | **76.92** | **86.27** | **41.59** | **36.78** |

---

[1] http://www.cs.cmu.edu/afs/cs.cmu.edu/project/theo-11/www/wwkb

## C. Ablation Study

Ablation experiments are raised to investigate whether the proposed methods are effective in improving the predictive power. To validate the benefits of the proposed structural feature generator, we reconstruct the way to capture structural information. Specifically, we use several aforementioned heuristic methods, including Common Neighbors, Resource Allocation and Jaccard to replace the previous structural feature generator and verify the effectiveness of obtained similarity score between neighbors. Note that "w/ CN" means our proposed method with Common Neighbors to compute similarity score which is previously introduced in the last chapter. "w/ RA" and "w/ Jaccard" also share the same meaning. The comparisons have been summarized in Figure 2. As the figure shows, our proposed framework (on the far right in the picture) performs better than any other NO-GAT with heuristic methods while capturing the structural information. Meanwhile these combination methods achieve better performance than GAT (on the far left in the picture) built by only attention mechanism without considering structural information.

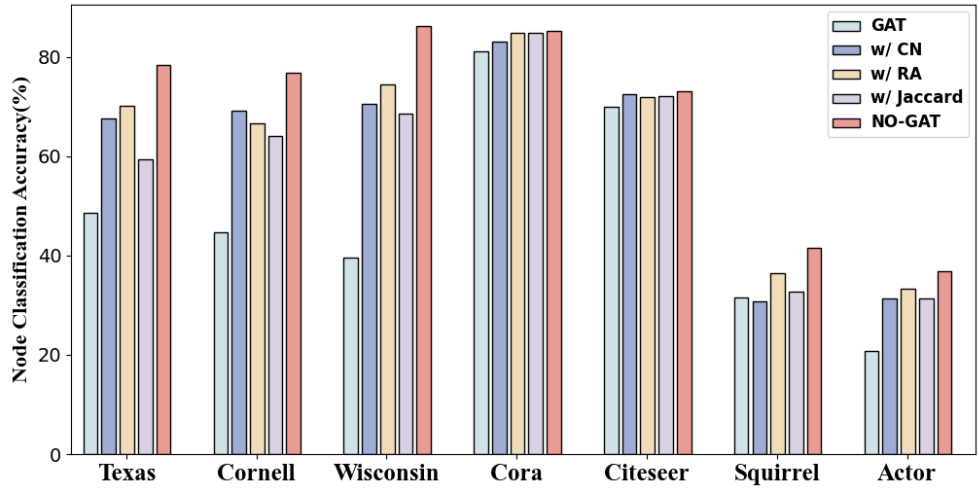

Figure 2. Ablation experiments on node classification of variant methods.

## VI. CONCLUSION

In this paper, we propose a novel and simple graph neural network framework named Neighbor Overlay-Induced Graph Attention Network for node classification. The key idea of NO-GAT is to capture both structural information and node features in a combined attention layer to compute appropriate weights for feature aggregation. Experiments demonstrate the effectiveness of our framework. In future work, we are about to further improve the performance of NO-GAT by enhancing the structure of graphs.